\documentclass[journal]{IEEEtran}
\usepackage{cite}
\usepackage{multirow}
\usepackage{cite}
\usepackage{amsmath,amssymb,amsfonts}
\usepackage{algorithmic}
\usepackage{graphicx}
\usepackage{float}
\usepackage{subfig}
\usepackage{textcomp}
\usepackage{colortbl} 
\usepackage{float}

\usepackage[table,svgnames]{xcolor}
\usepackage[justification=centering]{caption}
\usepackage{hyperref}
\usepackage{booktabs} 
\usepackage[T1]{fontenc}
\begin{document}

\title{WirelessGPT: A Generative Pre-trained Multi-task Learning Framework for Wireless Communication}
\author{\IEEEauthorblockN{Tingting Yang, Ping Zhang, Mengfan Zheng, Yuxuan Shi, Liwen Jing, Jianbo Huang, Nan Li} \\
\IEEEauthorblockA{\textit{Pengcheng Laboratory}, Shenzhen, China \\
		Emails: \{yangtt, zhangp02, zhengmf01, shiyx01, jinglw, huangjb, lin02\}@pcl.ac.cn}}

\maketitle

\begin{abstract}
This paper introduces WirelessGPT, a pioneering foundation model specifically designed for multi-task learning in wireless communication and sensing. Specifically, WirelessGPT leverages large-scale wireless channel datasets for unsupervised pretraining and extracting universal channel representations, which captures complex spatiotemporal dependencies. In fact, this task-agnostic design adapts WirelessGPT seamlessly to a wide range of downstream tasks, using a unified representation with minimal fine-tuning. By unifying communication and sensing functionalities, WirelessGPT addresses the limitations of task-specific models, offering a scalable and efficient solution for integrated sensing and communication (ISAC). With an initial parameter size of around 80 million, WirelessGPT demonstrates significant improvements over conventional methods and smaller AI models, reducing reliance on large-scale labeled data. As the first foundation model capable of supporting diverse tasks across different domains, WirelessGPT establishes a new benchmark, paving the way for future advancements in multi-task wireless systems.

\end{abstract}


\section{Introduction}
The rapid evolution of wireless communication systems has led to a growing demand for efficient and versatile models capable of supporting multiple tasks within a unified framework. In modern wireless networks, tasks such as channel modeling, signal prediction, environmental reconstruction, and target tracking are integral to ensuring reliable and efficient communication. Traditionally, these tasks have been approached independently, with separate methodologies tailored to each specific application. However, the increasing complexity of communication environments, driven by dynamic user behaviors and diverse deployment scenarios, necessitates the development of multi-task models that can simultaneously address these challenges within a cohesive framework. This need for integrated solutions has become particularly critical in the context of integrated sensing and communication (ISAC) systems, which aim to harmonize communication and sensing functions for enhanced network performance.

\begin{figure}[htbp]
\centerline{\includegraphics[width=0.5\textwidth,height=0.4\textwidth]{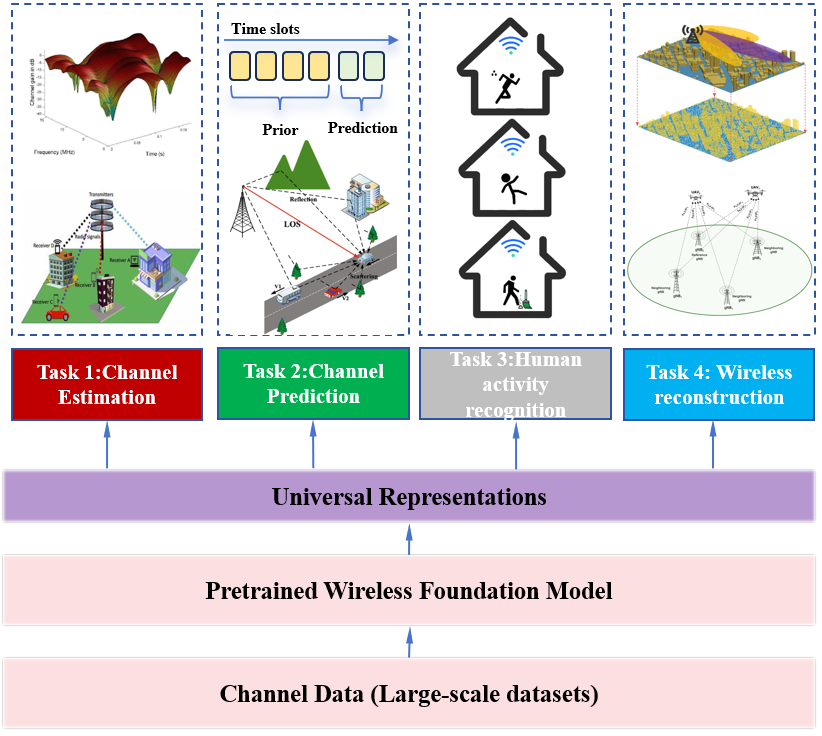}}
\caption{Framework of WirelessGPT}
\label{Pipeline}
\end{figure}
Despite significant advancements in wireless communication research, current methodologies face notable challenges. On the one hand, conventional models often rely on task-specific algorithms that lack flexibility and require substantial computational resources for each task. However, recent progress in deep learning has introduced models based on artificial intelligence (AI) capable of improving performance in a wide range of wireless communication tasks. However, these learnable models still face two main limitations as follows. First, they are typically specialized and fail to generalize effectively across spatial, temporal, and frequency domains,  hence fail to generalize across multiple tasks. Second, the scarcity of labeled data in many wireless scenarios poses a major obstacle that limits the applicability of supervised learning approaches. The above reasons have led to the fact that while large-scale foundation models have demonstrated remarkable success in fields such as natural language processing and computer vision, their application to wireless communication remains underexplored. Consequently, the lack of a unified framework capable of handling various communication and sensing tasks simultaneously has become a significant bottleneck in the advancement of ISAC technologies.
\renewcommand{\arraystretch}{1.5}
\begin{table*}[htbp]
    \centering
        \caption{Comparison among recent LLM-based models across different domains}
    \label{LLMComp}
	\begin{tabular}{m{4cm}<{\centering}|m{6cm}<{\centering}|m{1.5cm}<{\centering}|m{1.5cm}<{\centering}|m{1.5cm}<{\centering}}
		\hline
		\rowcolor[gray]{0.9} 
		\centering Model &  Tasks & Temporal& Spatial & Frequency \\
		\hline
		\centering LWM\cite{LWM} & Beam prediction; LoS/NLoS &  & \checkmark & \checkmark \\
		\hline
		\centering LLM4CP\cite{Liu_Liu_Gao_Cheng_Yang_2024}& Channel prediction & \checkmark& & \checkmark \\
		\hline
        \centering RFM\cite{ott2024radio}& Indoor localization & \checkmark & \checkmark & \\
		\hline
        \centering ChannelGPT\cite{ChannelGPT}& Channel prediction; CSI-Feedback & \checkmark & & \checkmark \\
		\hline
        \centering Trans4CP\cite{Trans4PC}& Channel Prediction & \checkmark & \checkmark &  \\
		\hline
        \centering CSI-GPT\cite{CSI-GPT}& CSI-Feedback & \checkmark &  & \checkmark \\
		\hline
         \centering CSI-based DT\cite{jiao20246G}& LoS/NLoS; UE position & & \checkmark & \checkmark \\
		\hline
        \centering WirelessGPT(ours))& Channel estimation; Channel Prediction; Poses recognition; Environment reconstruction & \checkmark& \checkmark & \checkmark \\
		\hline
	\end{tabular}

\end{table*}

In this paper, we propose WirelessGPT, a foundation model designed specifically for multi-task wireless communication and sensing. By leveraging large-scale wireless channel data for unsupervised pretraining, WirelessGPT is capable of learning complex multi-domain features, enabling it to generalize across a wide range of tasks with minimal fine-tuning. This approach addresses the aforementioned limitations by offering a unified framework that supports both communication and sensing applications, thereby enhancing system performance in data-limited scenarios. WirelessGPT aims to bridge the gap between existing task-specific models and the growing demand for integrated, multi-functional solutions in the wireless communication domain. Through this work, we aim to contribute a scalable and flexible model that advances the state of the art in ISAC research and lays the groundwork for future multi-task systems. The key contributions of this work are listed in the following.
\begin{itemize}
	\item \textbf{Innovative multi-dimensional correlation capture:} WirelessGPT introduces a novel pretraining approach that effectively captures spatial, temporal, and frequency correlations inherent in wireless channel data. This ensures comprehensive representation learning for diverse communication tasks.
	\item \textbf{Efficient channel representations generation:} By leveraging pretraining, WirelessGPT generates channel representations that maximize the utilization of cross-dimensional correlations, enabling efficient and robust execution of downstream tasks, including channel prediction, environmental reconstruction, and localization. These representations extends its applications beyond traditional communication tasks to include sensing functionalities, establishing itself as a unified foundation model for ISAC.
	\item \textbf{Scalability in Model Design:} The parameter size of WirelessGPT is significantly scalable, evolving from an initial size of 80M to a future design with 800M parameters. This scalability enhances its capacity to handle increasingly complex communication and sensing tasks.
	\item \textbf{A large-scale wireless channel dataset:} This work presents \textbf{Traciverse}, a self-developed and publicly available wireless channel dataset. This dataset, approximately 300 GB in size,  encompasses over 27 cities worldwide and more than 100 diverse scenarios, including dense urban areas, residential districts, urban centers, and suburban environments.
\end{itemize}

\section{Literature Review}

The increasing complexity and diversity of wireless communication systems have driven the need for models capable of performing multiple tasks, such as channel estimation, beamforming, and interference management, within a unified framework. Traditional task-specific approaches, while effective in isolated scenarios, struggle to generalize across diverse environments and often require large amounts of labeled data. This has motivated the exploration of task-agnostic foundation models for multi-task learning in communication systems.


Lee et al. proposed a framework that integrates pre-trained language models with physical layer communications, enhancing robustness under noisy conditions \cite{Ju-HyungLee}. However, its application is limited to single-task scenarios, lacking generalization across diverse tasks. Alikhani et al. introduced the Large Wireless Model (LWM), a task-agnostic foundation model leveraging Transformer-based architecture and self-supervised pretraining on large-scale wireless datasets \cite{LWM}. While LWM improves performance in multiple tasks such as channel modeling and beamforming, it focuses primarily on physical layer tasks, with limited applicability to broader multi-task ISAC scenarios. 

Other studies have explored multi-modal and multi-task learning paradigms. Jiao et al. \cite{jiao20246G} proposed a 6G-oriented framework employing contrastive learning to align channel state information (CSI) with environmental descriptions, enabling zero-shot learning and task-oriented fine-tuning for multiple tasks. Their approach highlights the generalization potential of foundation models across base station environments and tasks. Aboulfotouh et al. \cite{aboulfotouh2024building} introduced a Vision Transformer (ViT)-based radio foundation model, leveraging masked spectrogram modeling for pretraining. Their approach demonstrated enhanced generalization and reduced training costs for tasks like human activity sensing and spectrogram segmentation.

In parallel, Yu et al. introduced ChannelGPT \cite{ChannelGPT}, a large model designed to generate digital twin channels for 6G environment intelligence, emphasizing the integration of multimodal data for accurate channel parameter generation and scenario adaptation. Tian et al. \cite{tian2023multimodal} explored multimodal transformers for beam prediction by fusing data from diverse sensors like cameras, LiDAR, and GPS. Although their framework supports feature fusion, it primarily addresses beam prediction rather than a unified multi-task approach.

Ott et al. \cite{ott2024radio} demonstrated the potential of self-supervised learning for localization tasks using 5G channel measurements. Their transformer-based framework pre-trains models on unlabeled CIR data, showing strong performance in downstream localization tasks but limited applicability to other domains. Zayat et al. \cite{zayat2023transformer} introduced transformer-masked autoencoders (TMAE) for next-generation wireless networks, highlighting their potential in addressing challenges like channel estimation and resource allocation. However, their application remains focused on specific tasks rather than a comprehensive multi-task framework. Besides, a comprehensive comparison among recent LLM-based models is presented in Tab. \ref{LLMComp}.

Building upon these advancements, this study proposes WirelessGPT, a foundation model specifically designed for multi-task learning in integrated sensing and communication (ISAC) systems. By pretraining on large-scale wireless channel datasets and fine-tuning for diverse tasks, WirelessGPT addresses the limitations of existing models, offering a scalable and unified solution to enhance performance across a wide range of communication and sensing tasks. This work bridges the gap between task-specific models and the growing need for multi-task capabilities in next-generation wireless networks.

\section{Framework}
In this section, we introduce the datasets, methodologies and utilization pipeline for WirelessGPT, a pretrained foundation model for multi-task wireless communication and sensing.

\subsection{Dataset for Pretraining}\label{sec:data}
In our study, we leveraged three key datasets, Traciverse, SionnaRT\cite{sionna} and DeepMIMO\cite{alkhateeb2019deepmimo} to pretrain WirelessGPT. Among these, Traciverse is a self-developed and large-scale wireless channel dataset, and all three datasets are incorporated into the pretraining process to generate effective channel representations, for improve the downstream performance. The details of these datasets are provided as follows.

\textbf{Traciverse}: Traciverse is a self-developed and publicly available wireless channel dataset. The dataset encompasses over 27 cities worldwide and more than 100 diverse scenarios, including dense urban areas, residential districts, urban centers, and suburban environments. It is generated using the NVIDIA Sionna platform, enabling high-fidelity physical-layer link-level signal acquisition, capturing key propagation phenomena such as reflection, refraction, and scattering. The dataset supports flexible configurations of base station placement and density to accommodate different experimental settings. Furthermore, it spans multiple 6G candidate frequency bands (FR1, FR2, and FR3), making it a robust foundation for the development of WirelessGPT. The total dataset size is 300GB.

\textbf{SionnaRT}: This framework performs ray tracing to generate multipath wireless channels, capturing key propagation phenomena such as reflection, scattering, and diffraction. To support realistic simulations, SionnaRT utilizes high-fidelity scene maps from \textbf{Sensiverse}, which provides 25 diverse urban and rural environments. 

\textbf{DeepMIMO}: DeepMIMO is generated using ray-tracing data from Remcom's Wireless InSite, ensuring realistic propagation characteristics influenced by environmental geometry, materials, and transmitter-receiver configurations. This dataset is
highly flexible, allowing researchers to customize scenarios
by adjusting system and channel parameters, which makes it ideal for tasks covering a broad range of wireless communication applications.

\begin{figure}[tbp]
\centerline{\includegraphics[width=0.49\textwidth]{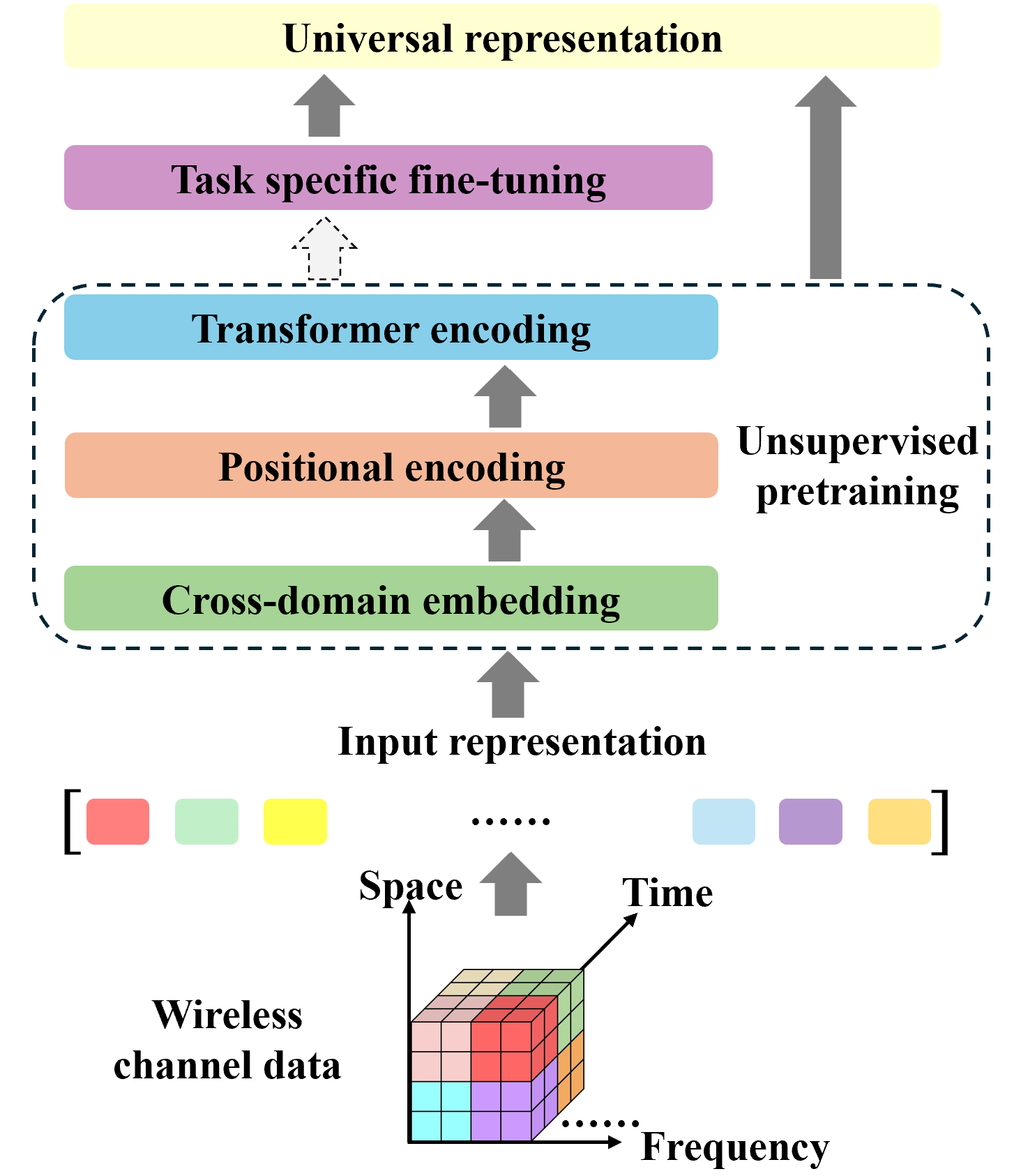}}
\caption{Framework illustration of pretraining }
\label{WGPT}
\end{figure}

\subsection{Model Pretraining}
To effectively model wireless communication channels, we propose a Transformer-based foundation model. Herein we detail the pretraining procedure of WirelessGPT, which contains data preprocessing, cross-domain representations, positional Encoding and optional fine-tuning.

\subsubsection{Data preprocessing}
Wireless signals inherently contain information across three dimensions: time, space, and frequency. The temporal dimension is represented by sampled snapshots, the spatial dimension captures characteristics such as antenna position, angle, and polarization, and the frequency dimension corresponds to subcarriers. To harness these structured features, the wireless signals are preprocessed into uniform three-dimensional slices before being fed into the model. After that, the fundation model will capture the cross-domain correlation among the partitioned patches. The reason why we utilize the Transformer architecture can be attributed to its ability to efficiently capture long-range dependencies and model complex interactions across different dimensions.

\subsubsection{Cross-domain representations}

The core mechanism of WirelessGPT relies on a multi-domain Transformer encoder, which extracts features from different domains. Specifically, spatial embeddings are generated through patch-based partitioning preserving geometric relationships. Temporal dependencies are captured using causal self-attention, modeling variations over time. Frequency-domain representations are obtained via spectral transformations, such as Fourier or wavelet transforms, encoding frequency-specific features. Meanwhile, positional encodings are implemented in three domains (note that positional encoding in the temporal domain is unnecessary), and are used to transform the embedding information into an efficient sequential form for input into the encoder. By integrating these embeddings through a cross-domain self-attention mechanism, WirelessGPT effectively learns joint representations, capturing intricate cross-domain dependencies, which enables improved performance across diverse wireless applications, demonstrating its adaptability to real-world scenarios.

\subsubsection{Scalable model design}
WirelessGPT offers a highly scalable model size, ranging from around 600K to 800M parameters, making it adaptable to a wide variety of wireless communication tasks with varying complexity and computational resources. This scalability allows the model to be efficiently deployed in diverse environments, from resource-constrained devices to high-performance systems. By adjusting the number of layers, attention heads, and hidden units, WirelessGPT can strike a balance between computational efficiency and model performance, ensuring that it can handle both simple tasks and more demanding scenarios. The ability to scale the model size enables WirelessGPT to improve its performance on tasks such as channel estimation, interference management, and signal reconstruction, without compromising on speed or accuracy. Furthermore, this flexibility makes WirelessGPT suitable for a broad range of applications, from mobile devices with limited processing power to large-scale base stations with ample computational resources. 

\subsubsection{Pretraining Process}

The pretraining phase follows an unsupervised learning approach, where some of the patches will be randomly masked, and the model is tasked with reconstructing the missing information. This strategy enables the model to learn meaningful representations by leveraging contextual information from the unmasked regions, meanwhile capturing the cross-domain dependencies. The dataset used for pretraining includes large-scale datasets such as Traciverse, DeepMIMO and SionnaRT, comprising millions of patches, ensuring diverse and comprehensive coverage of real-world channel conditions.

\subsection{Model Utilization Pipeline}
The application of our foundation model follows a structured pipeline to maximize performance across various downstream tasks:
\subsubsection{Pretraining} The model undergoes extensive pretraining on large-scale wireless datasets to learn universal representations of wireless channels.
\subsubsection{Universal Representation Generation} Using data from a specific downstream task, the pretrained model generates a universal representation, capturing essential signal features.
\subsubsection{Downstream Task Execution} The generated representation is utilized to perform the specific downstream task, improving accuracy and robustness.
\subsubsection{Fine-Tuning (Optional)}: For tasks requiring exceptionally high accuracy, the model can be fine-tuned using task-specific data. This iterative process involves reapplying steps 2 and 3 after fine-tuning, further optimizing performance. In the downstream tasks of WirelessGPT, this fine-tuning module is applied for tasks using a regression loss function, while is absent for classification tasks. Therefore, we label it with a dashed arrow.

By leveraging the foundation model to generate universal representations, we enhance the efficiency and accuracy of downstream tasks. The benefits of this approach will be further demonstrated in the experimental results section.

\section{Experiment}\label{sec:exp}
To demonstrate the generalization capability and performance superiority of WirelessGPT, it is adapted to a wide range of communication and sensing tasks, including channel estimation, channel prediction, and human activity recognition. Notably, in the channel prediction task, our foundational model achieves performance comparable to state-of-the-art (SOTA) framework based on LLM, particularly excelling in low-SNR regions.
\renewcommand{\arraystretch}{1.5}

\begin{table*}[h!]
\centering
\begin{tabular}{@{}p{4cm} p{4cm} p{4cm} p{4cm}@{}}
\toprule
 & \textbf{Channel estimation} & \textbf{Channel prediction} & \textbf{Human activity recoginition} \\ 
\midrule

\rowcolor{gray!20} \textbf{Environmental Details} &  &  &  \\
\#Subcarraiers        & 32       & 32            & 114 \\ 
\#Transmitt antenna             & 4       & 4$\times$4 (dual polarization)           & 3 \\ 
\#Receive antenna              & 4       & 1           & 3 \\ 
Center frequency & 2.4 GHz& 2.4 GHz & 5 GHz\\
Pilot length & 64 & --&--\\
Datasets(train/val/test) & WINNER II(6w/1w/2w) & QuadriGa(16w/3.2w/12.8w) &NTU-Fi HAR(800/200/200)\\
\rowcolor{gray!20} \textbf{Training Details Parameters} &  &  &  \\
Learning Rate         & 1e-4      & 1e-4            & 1e-3 \\ 
Batch size      & 512       & 512           & 16 \\ 
Optimizer   & Adam(0.9,0.999)       & Adam(0.9,0.999)            & Adam(0.9,0.999) \\ 
\rowcolor{gray!20} \textbf{Model Details} & \textbf{ResCNN} & \textbf{LSTM}  & \textbf{ResCNN} \\ \hline
Model parameters         & 161.5K       &  1.2M           & 316.1K \\ 
Structure      &  Conv1d (3×3, stride=1)       & --          & Conv2d (1→32, 3×3, stride=1)\\ 
\#blocks/layers & 4& 2&3 \\
feature length& 256       & 256            & 64 \\ 
\rowcolor{gray!20}  &\textbf{Transformer} & \textbf{Transformer}  & \\ \hline
Model parameters        & 1.3M      & 1.7M            &  \\ 
Structure & Enc-only       & Enc-Dec            &\\ 
\#enc/dec layers      & 2       & 2            &  \\ 
\#attention heads   & 4       & 4            &  \\ 
hidden/forward dimension &256/512 &256/512 &\\
\bottomrule
\end{tabular}
\caption{Implementation details of WirelessGPT for downstream tasks}
\label{tab:parameters}
\end{table*}

\subsection{Channel Estimation}

Channel estimation is a fundamental task in wireless communication systems, which plays a critical role in ensuring reliable data transmission over multipath and time-varying fading channels. It involves estimating the characteristics of the communication channel, such as the amplitude, phase, and impulse response, based on prior knowledge of the transmitted signal and the received signal. In this experiment, we consider an OFDM system with 32 sub-carriers and a $32$ antenna array. The wireless channels are simulated using the Wireless World Initiative New Radio (WINNER) II channel model\cite{8045088}. Additionally, a 64-length pilot sequence and QPSK modulation are employed to complete the channel estimation, while average SNR is sampled from -5 to 10.

\begin{figure}[tbp]
\centerline{\includegraphics[width=0.45\textwidth]{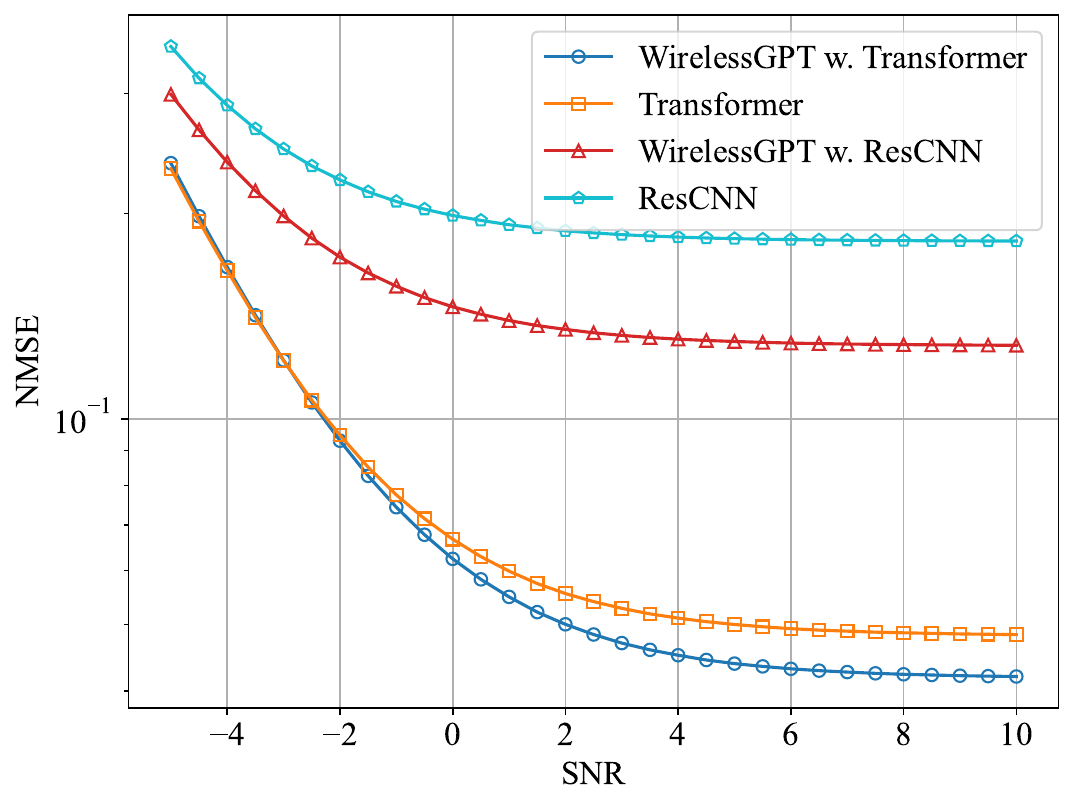}}
\caption{NMSE comparisons between WirelessGPT and baseline models for channel estimation}
\label{Estimation_SNRvsNMSE}
\end{figure}

For performance evaluation, universal representations and raw channel data are fed into two simple downstream models for comparison. These models are denoted as WirelessGPT w. [model name] and [model name], where the model name is selected from Transformer and Residual Convolutional Neural Network (ResCNN). Detailed configurations of the downstream models are provided in Table \ref{tab:parameters}. The normalized mean square error (NMSE) is utilized to quantify the estimation error for each model. As shown in Fig. \ref{Estimation_SNRvsNMSE}, we present the NMSE performance of different baselines for the channel estimation task. Overall, WirelessGPT demonstrates superior performance compared to its competitors, attributed to the efficient channel representations extracted by the pre-trained foundational model. Specifically, compared to the raw channel materials, the preprocessed representations result in a \textbf{$14.09\%$ NMSE reduction for the Transformer} and a \textbf{$41.44\% $ NMSE degradation for ResCNN}. This difference can be attributed to that the channel representations can effectively expand the receptive field after capturing the temporal-spatial dependency, and consequently obtain a larger performance promotion in the CNN-based architecture.

Furthermore, compared to the raw channel inputs, the efficient channel representations also contribute to reducing the training and inference time of downstream models. This is because the pre-trained foundation model processes the complex channel matrix into a more structured and easily interpretable format for downstream models. As shown in Table. \ref{Est}, WirelessGPT shows both training and inference complexity degradation in estimation task, when compared to the baselines.

\renewcommand{\arraystretch}{1.5}
\begin{table}[htbp]
	\centering
	\caption{Comparisons of parameters and time complexities for estimation task}\label{Est}
	\begin{tabular}{m{1.2cm}<{\centering}|m{1.5cm}<{\centering}|m{1.3cm}<{\centering}|m{1.5cm}<{\centering}|m{0.9cm}<{\centering}}
		\hline
		\rowcolor[gray]{0.9} 
		\centering &  WirelessGPT w. Transformer & Transformer& WirelessGPT w. ResCNN & ResCNN \\
		\hline
		\centering Parameters (M) & 79.6+1.7 & 1.7 & 79.6+ 0.9 & 0.9 \\
		\hline
		\centering Training time (ms)& 188.3 & 217.1& 115.2 &129.4\\
		\hline
        \centering Inference time (ms)& 2.31 & 2.48 & 1.53 & 1.98\\
		\hline
	\end{tabular}
\end{table}
\subsection{Channel Prediction}
\begin{figure*}[htbp]
\subfloat[]{
			\includegraphics[width=0.487\textwidth]{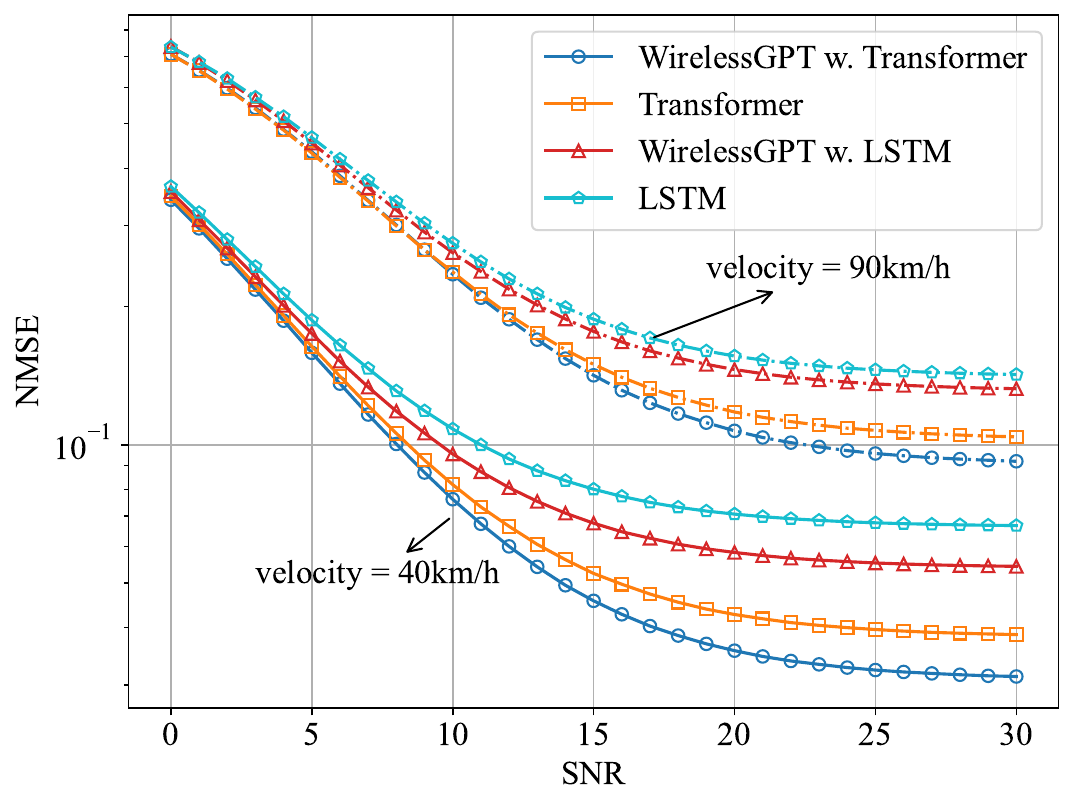}
			\label{WGPTvsBaseline}}
\subfloat[]{
			\includegraphics[width=0.48\textwidth]{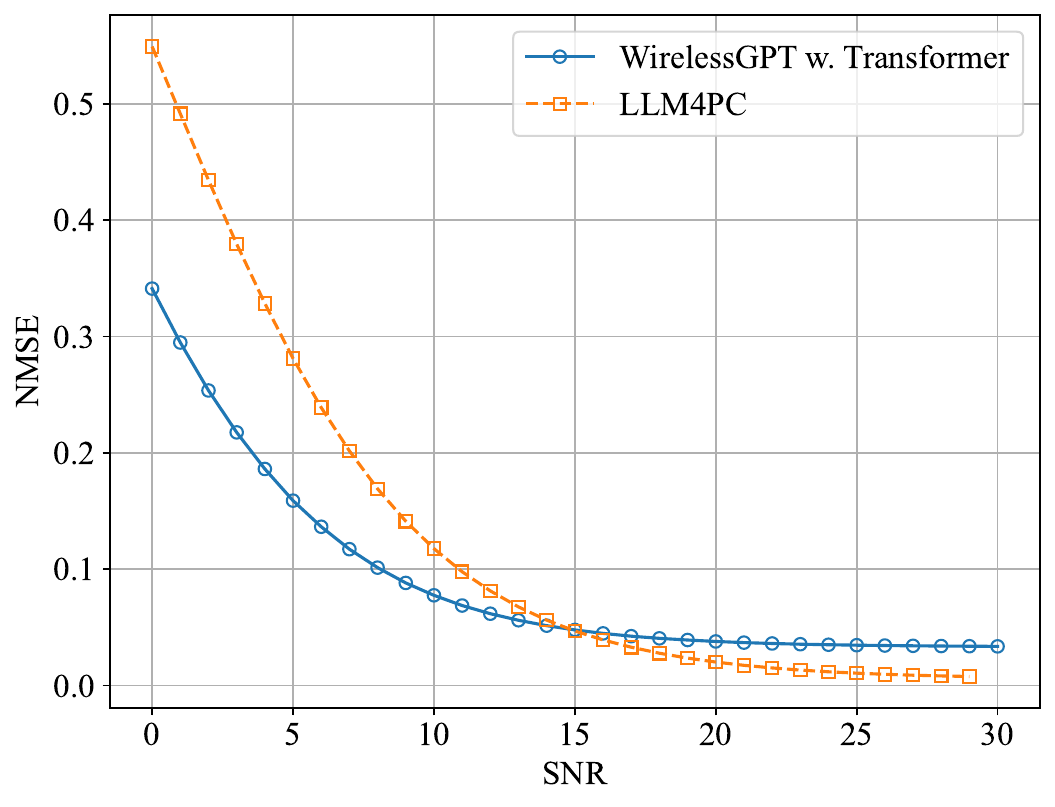}
			\label{WGPTvsLLM4CP}}
\caption{NMSE comparisons against channel quality for channel prediction a) WirelessGPT vs Transformer/LSTM b) WirelessGPT vs large language model}
\label{Prediction_SNRvsNMSE}
\end{figure*}

As a task closely related to channel estimation, channel prediction focuses on forecasting the future state of the wireless channel based on its estimated historical behavior. This task is critical in wireless communications, enabling proactive resource allocation and adaptive system optimization. In this experiment, we consider a MISO-OFDM system equipped with a $4\times4$ antenna array. The temporal-dynamic wireless channels are generated using QuaDRiGa \cite{QuadriGa}, a widely adopted framework for simulating time-varying channel environments compliant with 3GPP standards. The uplink channel operates at a center frequency of 2.4 GHz with 32 subcarriers. In this scenario, the task involves predicting the channel state information (CSI) for four future slots based on the historical data of 16 slots. The average SNR is sampled within the range of 0 to 30 dB, and the user velocity is varied between 40 and 100 km/h to simulate diverse and dynamic communication conditions. This setup allows for evaluating the model's performance under realistic and challenging temporal dynamics.

To align with the objective of sequence prediction, we adjusted the selection of downstream models, namely model name can be selected between Transformer and Long Short-Term Memory (LSTM). In Fig. \eqref{WGPTvsBaseline}, NMSE behaviors against SNR among competitors are plotted by fixing velocity equals to 40 and 90 km/h. Evidently, the performance gains brought by efficient channel representations in sequential predictions have also been demonstrated, which is attributed to the pre-trained foundation model's ability to model and represent the complex channel matrix across the spatial, temporal, and frequency dimensions.
\begin{table*}[htpb]
	\centering
	\caption{Comparisons of parameters and time complexities for prediction task}\label{complexity2}
	\begin{tabular}{m{3cm}<{\centering}|m{2.5cm}<{\centering}|m{2.5cm}<{\centering}|m{2.5cm}<{\centering}|m{2.5cm}<{\centering}|m{2.5cm}<{\centering}}
		\hline
		\rowcolor[gray]{0.9} 
		\centering &  WirelessGPT w. Transformer & Transformer& WirelessGPT w. LSTM & LSTM & LLM4CP \\
		\hline
		\centering Parameters (M) & 79.6+1.7 & 1.7 & 79.6+1.4 & 1.4 & 1.7+82.9\\
		\hline
		\centering Training time (ms)& 232.9 & 312.8& 104.8 & 192.5 &449.9\\
		\hline
        \centering Inference time (ms)& 2.26 & 2.56 & 5.12 & 7.08 & 17.23\\
		\hline
	\end{tabular}
\end{table*}

Fortunately, Liu et al. \cite{Liu_Liu_Gao_Cheng_Yang_2024} recently proposed an LLM-based framework for channel prediction, namely LLM4CP, enabling comparisons between WirelessGPT with foundation models of a similar scale. As shown in Fig. \eqref{WGPTvsLLM4CP}, WirelessGPT equipped with a 1.7M-sized Transformer as the downstream model, demonstrates comparable NMSE performance to LLM4CP, which is particularly designed for channel prediction. More specific, WirelessGPT achieves better NMSE performance in relatively poor channel conditions with SNR below 15, but falls short of LLM4CP when under pure channel conditions. However, the advantages of WirelessGPT are not limited to supporting different downstream tasks with the same foundation model, but also lie in its scalability—the performance bottleneck can be lifted with improvements in the downstream model. In other words, if we replace the downstream model with a more advanced channel prediction model (e.g., LLM4CP), we expect to achieve even better performance.

Finally, we also compared the training and inference times of various baseline schemes in Table \ref{complexity2}. It can be observed that, due to the efficient representation of channel data during the pre-training phase, WirelessGPT demonstrates an improvement in time complexity even compared to smaller models\footnote{The training and inference time of LLM4CP appear different since different GPU devices are used in comparison with \cite{Liu_Liu_Gao_Cheng_Yang_2024}.}.


\renewcommand{\arraystretch}{1.5}

\subsection{Human Activity Recognition}
Human activity recognition is the third task in our case study, demonstrating the generality of the foundation model WirelessGPT and its potential for compressing CSI signal information. This task involves classifying human activities using WiFi CSI amplitude measurements collected by three routers in an indoor environment. Six activities—running, walking, falling down, boxing, circling arms, and cleaning the floor—were investigated, with data collected from 20 subjects (13 males and 7 females). This dataset is collected by researchers described in \cite{yang2022efficientfi} and details are provided in Table \ref{tab:parameters}. 

Each measurement is first processed through sampling and reshaping to meet the input requirements of the foundation model. By performing multiple sampling operations, we preserve detailed information while transforming the data into a compact format. The processed input is then fed into the foundation model, where its encoder architecture generates a universal representation. This representation significantly compresses the original data while retaining essential features for downstream tasks. The compression is achieved through a multi-head attention mechanism, allowing the output representation to attend to all input patches. As a result, the representation is more compact and contains denser information while preserving all input features.

By directly using this universal representation as the input for a classifier implemented with a 2D CNN featuring residual connections and 316.1K parameters, the training loss converges within 3 minutes on a GeForce 3060 GPU, achieving a testing recognition accuracy of 98.11\%. This performance is comparable to state-of-the-art results. A comparison of performance between using raw data and the universal representation as inputs for the downstream task is presented in Table \ref{tab:raw_vs_embedding}.

\begin{table}[htpb]
	\label{tab:raw_vs_embedding}
	\centering
	\caption{Comparison of Performance Metrics Between Raw Measurement data and Universal Representation}
	\label{tab:raw_vs_embedding}
	\begin{tabular}{m{2.5cm}<{\centering}|m{2.5cm}<{\centering}|m{2cm}<{\centering}}
		\hline
            \rowcolor[gray]{0.9} 
		~ & ResCNN & WirelessGPT w. ResCNN \\ \hline
		Accuracy (\%) & 96.5 & 98.1 \\ \hline
		Input Dimension & 3×114×2000 & 72×64 \\ \hline
		Flops (G) & 210.39 & 1.42 \\ \hline
		Training Time (ms) & 151.9 & 9.9 \\ \hline
	\end{tabular}
\end{table}

The measurement data in this case is entirely new to the pre-trained foundation model. This dataset consists of indoor WiFi CSI measurements with only amplitude information, whereas the model was pre-trained using outdoor complex-valued CSI matrices between base stations and end users. This demonstrates the generality of the foundation model. Additionally, the universal representation extracted by the foundation model significantly compresses the CSI signal, reducing the data for the downstream classifier to just 0.67\% of the original measurement data.

\subsection{Wireless Reconstruction} The evolution of wireless networks toward integrated sensing and communication (ISAC) systems is paving the way for the next generation of communication technologies. These systems are designed to seamlessly integrate wireless communication with environmental sensing, enabling a wide range of applications, from precise mobile positioning and environmental monitoring to autonomous navigation. As wireless communication and sensing converge, reconstructing accurate and real-time models of the wireless environment is becoming increasingly critical. The ability to reconstruct these environments in a scalable and efficient manner can significantly enhance resource optimization, improve network performance, and support applications in autonomous systems, smart cities, and other advanced technological domains.

The experimental evaluation of environment reconstruction focused on both line-of-sight (LOS) and non-line-of-sight (NLOS) conditions, including first-order reflections simulated by setting the MaxNumReflections parameter in Sionna to 1. To ensure comprehensive testing, the experiments were conducted across diverse environments, including urban and rural scenarios, and spanned a wide range of frequencies, from low-frequency bands (3.5 GHz) to high-frequency millimeter-wave bands (10 GHz, 28 GHz, and 60 GHz). Different subcarrier spacings (SCS) were configured to match 5G NR standards, including 15 kHz for 3.5 GHz, 30 kHz for 10 GHz, and 120 kHz for 28 GHz and 60 GHz, ensuring compatibility with the channel characteristics at each frequency.

\subsubsection{Results} The experimental evaluation of environmental reconstruction highlights its effectiveness in reconstructing wireless environments with high accuracy. The training process demonstrates consistent convergence of the loss function, particularly the Chamfer Distance, indicating the model’s ability to effectively learn spatial and spectral relationships within the multimodal input data. The convergence also reflects the robustness of the training strategy, which incorporates mini-batch processing, positional encodings, and the Chamfer Distance as the reconstruction loss metric.

The simulated environment for evaluation included configurations where each base station was equipped with one omnidirectional transmitting antenna and 16 omnidirectional receiving antennas arranged in a 4 × 4 planar array. The inter-element spacing of the receiving antenna array was set to half the wavelength, based on a carrier frequency of 10 GHz. This configuration, combining omnidirectional antennas and a planar array, ensures comprehensive spatial coverage and captures diverse propagation characteristics within the environment. During the inference phase, WirelessGPT generated a 3D point cloud representation consisting of 250 scatterer points. These scatterer points correspond to strong scatterers, such as building edges and corners, highlighting the model's capability to extract critical features.

\subsubsection{Analysis} The experimental results demonstrate the effectiveness of WirelessGPT in reconstructing wireless environments with high accuracy and adaptability across diverse scenarios. The model achieves superior resolution in both low-frequency and millimeter-wave bands, addressing the challenges posed by complex multipath propagation and dynamic environments. By leveraging the generative capabilities of the Transformer architecture, WirelessGPT excels in capturing strong scattering features, such as building edges and corners, which are critical for accurate point cloud representations.

\section{Conclusion and Future Work}
Further efforts are directed toward scaling the model to larger parameter sizes to explore its adherence to scaling laws and to evaluate its broader applicability. As one of the early attempts to integrate communication and sensing tasks within a single model, WirelessGPT lays the groundwork for advancing ISAC technologies.

\section*{Acknowledgment}
This work is supported by the National Key R\&D Program of China (Grant No.2024YFE0200801, No.2024YFE0200804)

\small

\bibliography{arXivWirelessGPT}
\end{document}